\documentclass{article}
\usepackage{spconf,amsmath}
\usepackage{booktabs, tabularx, multirow, graphics, hhline, boldline}
\usepackage{array, amssymb}
\usepackage{enumitem}
\usepackage{graphicx, color}
\usepackage{cite}
\usepackage{xr}
\usepackage{hyperref}
\usepackage[normalem]{ulem}
\usepackage[table,xcdraw]{xcolor}

\newcommand{\fig}{Fig.}
\newcommand{\figs}{Figs.}

\newcommand{\sect}{Sec.}

\newcommand{\tuning}{{\it all-layers}}

\newcommand{\wavtovec}{{\it W2V2}}
\newcommand{\wavtovecl}{{\it W2V2-large}}
\newcommand{\wavtovecs}{{\it W2V2-small}}
\newcommand{\fastvgs}{{\it  FaST-VGS}}
\newcommand{\fastvgsp}{{\it  FaST-VGS+}}
\newcommand{\avhubert}{{\it AV-HuBERT}}
\newcommand{\hubert}{{\it HuBERT}}
\newcommand{\hubertl}{{\it HuBERT-large}}
\newcommand{\huberts}{{\it HuBERT-small}}
\newcommand{\xlsr}{{\it XLSR-53}}
\newcommand{\wavlm}{{\it WavLM}}

\DeclareMathOperator*{\argmax}{arg\,max}

\title{Comparative layer-wise analysis of self-supervised speech models}
\name{Ankita Pasad, Bowen Shi, Karen Livescu}
\address{Toyota Technological Institute at Chicago}
\begin{document}
\ninept
\maketitle
\begin{abstract}
\vspace{-.025in}
Many self-supervised speech models, varying in their pre-training objective, input modality, and pre-training data, have been proposed in the last few years.  Despite impressive successes on downstream tasks, we still have a limited understanding of the properties encoded by the models and the differences across models.  In this work, we examine the intermediate representations for a variety of recent models.  Specifically, we measure acoustic, phonetic, and word-level properties encoded in individual layers, using a lightweight analysis tool based on canonical correlation analysis (CCA). We find that these properties evolve across layers differently depending on the model, and the variations relate to the choice of pre-training objective. We further investigate the utility of our analyses for downstream tasks by comparing the property trends with performance on speech recognition and spoken language understanding tasks. We discover that CCA trends provide reliable guidance to choose layers of interest for downstream tasks and that single-layer performance often matches or improves upon using all layers, suggesting implications for more efficient use of pre-trained models.\footnote{Codebase: \href{https://github.com/ankitapasad/layerwise-analysis/}{https://github.com/ankitapasad/layerwise-analysis/}}  
\end{abstract}

\vspace{-.025in}
\section{Introduction}
\vspace{-.03in}
Self-supervised models have become a nearly ubiquitous approach for learning speech representations and improving performance on downstream tasks~\cite{mohamed2022self,borgholt2022brief,Yang2021SUPERBSP, shon2022slue, tsai2022superb}, but our understanding of their properties and strategies for their use is still limited.  Some recent work has begun developing an understanding of the extent of speaker~\cite{fan2020exploring, feng2022silence}, para-linguistic~\cite{shah2021all, li2022exploration}, articulatory~\cite{ji2022predicting}, and acoustic-linguistic~\cite{pasad2021layer} properties encoded in these models, which in some cases has resulted in improved fine-tuning strategies~\cite{pasad2021layer, feng2022silence}.

In this work, we contribute to understanding pre-trained models beyond their empirical performance. We extend our previous work that focused on wav2vec 2.0~\cite{pasad2021layer} to a range of additional models. We measure the extent of acoustic, phonetic, and word content encoded in individual layers for 11 pre-trained models, using a lightweight analysis tool based on canonical correlation analysis (CCA). We find that phonetic and word information concentrates in different layers for different models, and the layer-wise trends relate to the pre-training objectives, despite differences in training data.

We also seek to leverage these insights for downstream tasks.  In recent work, ``frozen" pre-trained representations are commonly used for a variety of tasks without fine-tuning~\cite{Yang2021SUPERBSP}.  We study whether high phone/word content according to our analysis also correlates with the usefulness of the corresponding frozen representation for related tasks.  We find that (i) both phone and word content are well-correlated with layer-wise speech recognition performance, (ii) word content is highly correlated with layer-wise performance on spoken language understanding (SLU) tasks, and (iii) performance from a single intermediate layer can match or outperform using all layers (frozen), which suggests avenues for the development of more efficient application of pre-trained models.

\vspace{-.1in}
\section{Background}
\label{sec:background}
\vspace{-.05in}
Self-supervised models are trained with an objective function that solves a pretext task formulated
using unlabeled data. Many self-supervised speech representation models have been proposed in the last few years~\cite{mohamed2022self}. We present analysis for {\it eleven speech models} differing in (i) training objective, (ii) training data modality (using either only speech or image-speech pairs), (iii) training data languages (English or multilingual), and (iv) model size. The pre-trained checkpoints for these models are obtained from their publicly available sources.

A typical model architecture has two components. The raw audio (or filter banks) is passed through a set of convolutional layers (or a linear projection). The resulting frame-level \emph{local} features are then processed through a set of self-attention layers. The models we use have 7 convolutional (or 1 linear) and 12 or 24 transformer layers, with the exception of \fastvgs~\cite{peng2022fast} (discussed below).

All the models in this work use a masking-based pretext task, thus using both left and right context to recover the masked segment (target). The target comes from either the local features -- wav2vec 2.0 (\wavtovec)~\cite{baevski2020wav2vec}, \xlsr~\cite{conneau2020unsupervised}, \fastvgsp~\cite{peng2022self} or from one of the intermediate transformer layers, represented as a discrete cluster ID -- \hubert\footnote{\wavtovec \ and \hubert \ small models are pre-trained on 960 hours LibriSpeech, and the corresponding large models on 60k hours LibriLight data.}~\cite{hsu2021hubert}, \wavlm\footnote{\wavlm {\it -small} is pre-trained on 960 hours LibriSpeech and \wavlm{\it -large} on 94k hours consisting of LibriLight, GigaSpeech, and VoxPopuli.}~\cite{chen2022wavlm}, \avhubert\footnote{\avhubert \ models are pre-trained on LRS3.}~\cite{shi2022robust}. Models of the first type are trained with a contrastive loss and the latter with a classification loss. 

In addition to discrete cluster ID prediction, \wavlm \ augments the input data to simulate noisy/overlapped speech. The classification loss for \wavlm \ uses the cluster IDs from \hubert's intermediate layers (the same layers that are used in \hubert's pre-training). \xlsr \ is trained on spoken data from 53 languages. For the audio-visual models, \avhubert, \fastvgs\footnote{\fastvgs \ and \fastvgsp \ models are pre-trained on SpokenCOCO.}\, and \fastvgsp, we use the audio branch alone, as our analyses use only speech input. 
The \fastvgs \ and \fastvgsp \ models are initialized from the pre-trained \wavtovecs \ model\footnote{\fastvgs \ uses 8 of the 12 \wavtovecs \ transformer layers; \fastvgsp \ uses all 12 layers.  The \wavtovecs \ CNN layer weights are kept frozen.} along with additional CNN, self-attention, and cross-attention layers. This ``audio branch" is trained along with a visual branch with a cross-modal contrastive loss.
\fastvgsp \ extends \fastvgs \ with an additional masking-based contrastive loss. The \avhubert \ model is trained on a lipreading dataset with a pre-training objective that uses multi-modal discrete units, and the audio branch computes local features using filter bank input passed through a single linear layer.
\vspace{-.1in}
\section{Analysis methods}
\vspace{-.1in}
As in our previous work analyzing \wavtovec~\cite{pasad2021layer}, we use CCA to measure the content of different acoustic and linguistic properties encoded in the pre-trained speech models. CCA~\cite{harold1936relations} is a statistical technique that measures the relationship between two continuous-valued random vectors as represented by the maximum correlations between their linear projections.

CCA takes as input $n$ pairs of vectors $\{(x_1, y_1), ..., (x_n, y_n)\}$, sampled from the random vectors (or ``views") $X\in\mathbb{R}^{d_1}, Y\in\mathbb{R}^{d_2}$, and returns {\it canonical correlations}, a correlation-based measure of similarity between the two views. The problem can be defined iteratively as follows: First, we define the directions of maximum correlation between linear projections of $X$ and $Y$: $ v_1, w_1 = \argmax_{v, w} \text{corr}(v^TX, w^TY) $.
The subsequent directions $v_i, w_i \; \forall \ i \in [2, \min(d_1, d_2)$] maximize the same correlation subject to each new projection being uncorrelated with others in the same view.

We use a variant of CCA, {\it projection-weighted CCA (PWCCA)}\cite{morcos2018insights}, which computes a weighted mean of the canonical correlations $\rho_i = \text{corr}(v_i^TX, w_i^TY)$, with higher weights for directions accounting for a higher proportion of the input. PWCCA has been found to be more robust to spurious correlations in the data.  To ensure the stability of the matrix inverse operations involved in the solution, two regularization parameters $\epsilon_x$ and $\epsilon_y$ are added to the covariance matrices. CCA similarity has a maximum value of 1.

We use PWCCA to measure the similarity between layer representations and various variables of interest. When comparing to discrete variables, the discrete label IDs are converted to one-hot vectors. As external variables of interest we use (80-dimensional) mel filter bank features ({\it CCA-mel}), phone labels ({\it CCA-phone}), and word labels ({\it CCA-word}) as ways to assess the local acoustic, phone identity, and word identity information encoded in the model representations respectively.

{\bf Implications for downstream tasks.} 
In addition to the CCA-based measures, we also measure the performance of individual layers on downstream tasks, using a prediction model on top of the frozen representations, trained on labeled data for the task. See \sect~\ref{sec:exp-details} for task-specific details.

We seek to measure the utility of our CCA-based analysis measures for downstream tasks. Specifically, {\it do layers with high property content, as measured by our analysis, also perform better on downstream tasks that are expected to benefit from that property?} We report Spearman's $\rho$ rank correlation between the layer-wise analysis scores (CCA-phone, CCA-word) and the performance of individual layers on downstream tasks. We choose \wavtovec \ and \hubert \ models for these experiments since these have distinct objectives and all other models we analyze are derived from these two.

\vspace{-.15in}
\section{Experiment details}
\vspace{-.12in}
\label{sec:exp-details}
We analyze the representations extracted from each layer of the models described in \sect~\ref{sec:background}. In addition, we also extract representations from a randomly initialized model ({\it rand-init}) with 7 CNN and 12/24 transformer layers, in order to check whether the model architecture itself has an inductive bias that could account for the observed trends.

We extract frame-level, phone-level, and word-level representations as in~\cite{pasad2021layer}. For frame-level analyses, 500 utterances ($\sim$180k frames) are sampled from the LibriSpeech~\cite{panayotov2015librispeech} dev-clean set. The phone-level and word-level representations are extracted from $\sim$7k instances (segments) each, covering 39 phones and 500 words. Three sets of such samples are generated. To avoid overfitting, a sample set is partitioned into ten splits, of which eight are used for training (learning the projection matrices $V$ and $W$), one for hyperparameter tuning ($\epsilon_x$ and $\epsilon_y$), and the last for testing. For robustness, this process is repeated three times with a different train-dev-test split each time. The reported CCA similarity for each layer is therefore an average of nine correlation scores as each of the three sample sets is used three times with a different test set each time.\footnote{We note that this setup is different from our previous work~\cite{pasad2021layer} where we don't split the data and in effect report scores on the training set. The trends remain similar in either setup, but the approach here is more robust to potential overfitting.}

For the experiments on downstream tasks, we consider tasks related to phonetic, word-level, and semantic content, specifically: phonetic recognition (PR), automatic speech recognition (ASR), and intent classification (IC) using the same prediction model and data as SUPERB~\cite{Yang2021SUPERBSP}; and the action and scenario classification tasks from the Spoken Language Understanding Resource Package (SLURP-action, SLURP-scenario)~\cite{Bastianelli2020SLURPAS}.  For the SLURP tasks, we use the same downstream classifier architecture as for the IC task of SUPERB~\cite{Yang2021SUPERBSP}.  We report layer-wise performance using individual layer representations as input to the task-specific classifier. For the large models, we only evaluate every other layer to limit computation cost. The ASR model is evaluated without a language model. For these experiments, we use the s3prl toolkit.\footnote{\href{https://github.com/s3prl/s3prl}{https://github.com/s3prl/s3prl}}
\vspace{-.1in}
\section{Findings}
\vspace{-.1in}
We first present the results of our analysis of multiple models using CCA similarity (Sec.~\ref{sec:findings1},~\ref{sec:findings2}) and then relate these results to performance on downstream tasks (Sec.~\ref{sec:corr}). We refer to the output of transformer layer $l$ as the representation at layer $l$ and the output of the CNN feature encoder (or linear layer for \avhubert) as layer 0 or ``local features".
\begin{figure}[tb]
\begin{minipage}[b]{1.0\linewidth}
\small

 \centering
 \centerline{\includegraphics[width=8cm, trim=0 184 0 0, clip]{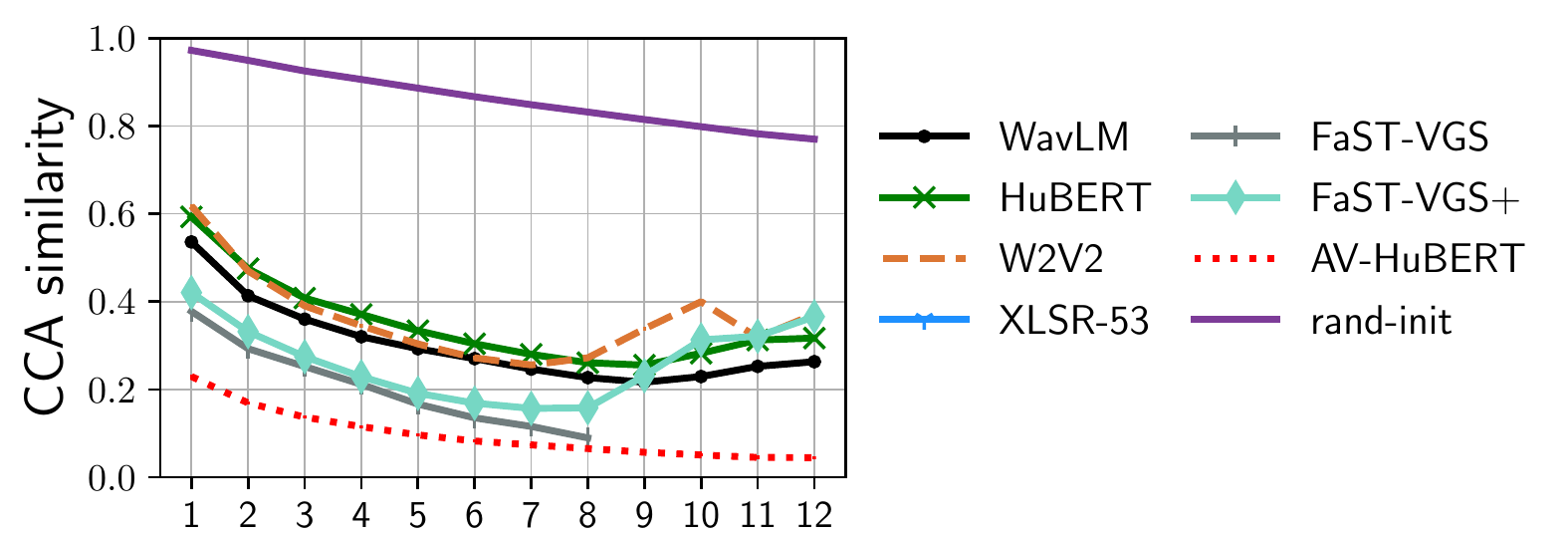}}
\end{minipage}
\begin{minipage}[b]{1.0\linewidth}

\vspace{-0.05cm}
\footnotesize
 \centering
 \centerline{\includegraphics[width=8cm, trim=0 0 0 155, clip]{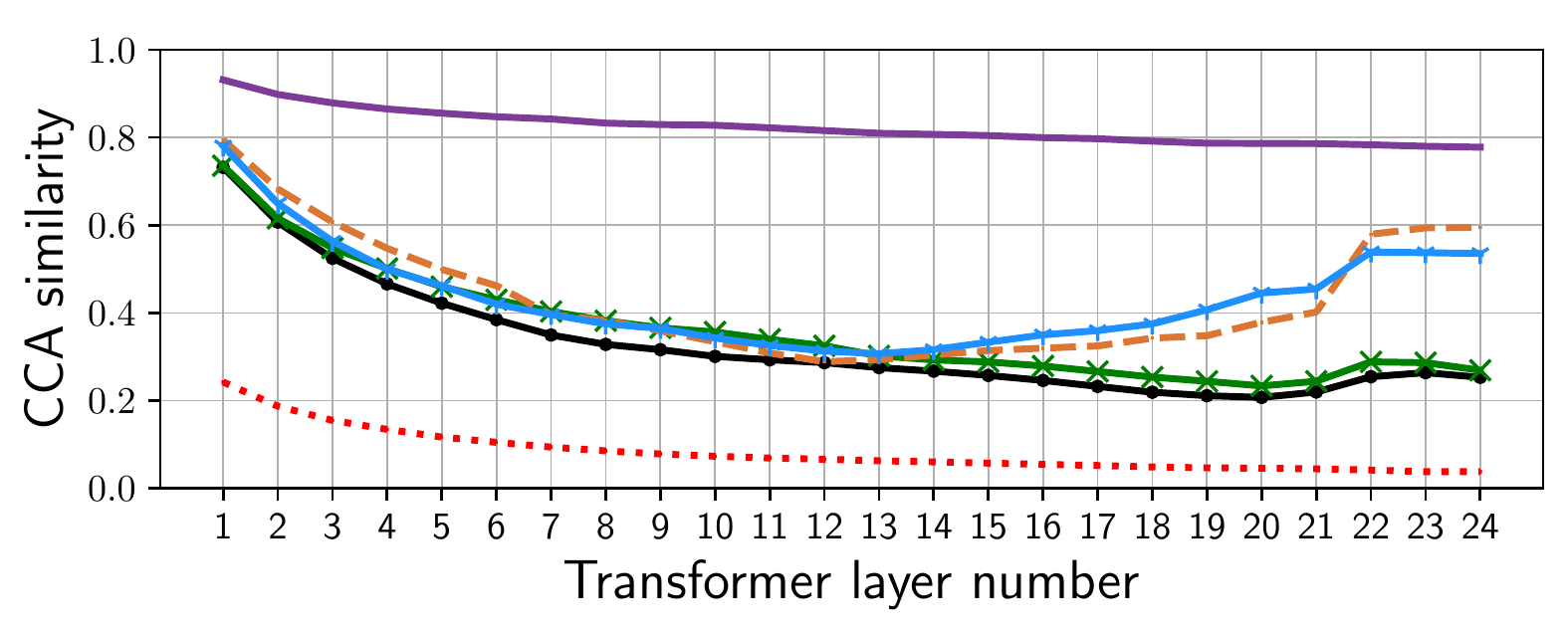}}
\end{minipage}

\vspace{-0.3cm}
\caption{\it CCA similarity with local features.
}
  \label{fig:cca-intra}
\vspace{-0.6cm}
\end{figure}
\vspace{-.1in}
\subsection{Evolution of representations across layers}
\label{sec:findings1}
\vspace{-.08in}
Fig.~\ref{fig:cca-intra} shows how the representations in each transformer layer relate to the input ``local features". Some models have a clear autoencoder-style pattern, i.e., high similarity with the input for the initial and final layers and a drop in similarity for the middle layers.  This behavior is most prominent in models trained to recover (in some sense) the local features (\wavtovec, \xlsr, and \fastvgsp) and less prominent in models that recover an intermediate transformer layer (\wavlm, \hubert, and \avhubert). Audio-visual models (\avhubert, \fastvgs, and \fastvgsp) diverge the most from the input as we go deeper into the network, possibly learning more from the visual modality and retaining less from the audio modality. 

In the randomly initialized model, the representations change little across layers, providing evidence that the findings for other models are due to the effect of pre-training and not an artifact of the architecture.

\vspace{-.12in}
\subsection{Measures of acoustic/linguistic content}
\label{sec:findings2}
\vspace{-.08in}

\begin{figure}[tb]
\begin{minipage}[b]{1.0\linewidth}
\small

 \centering
 \centerline{\includegraphics[width=8cm, trim=0 180 0 0, clip]{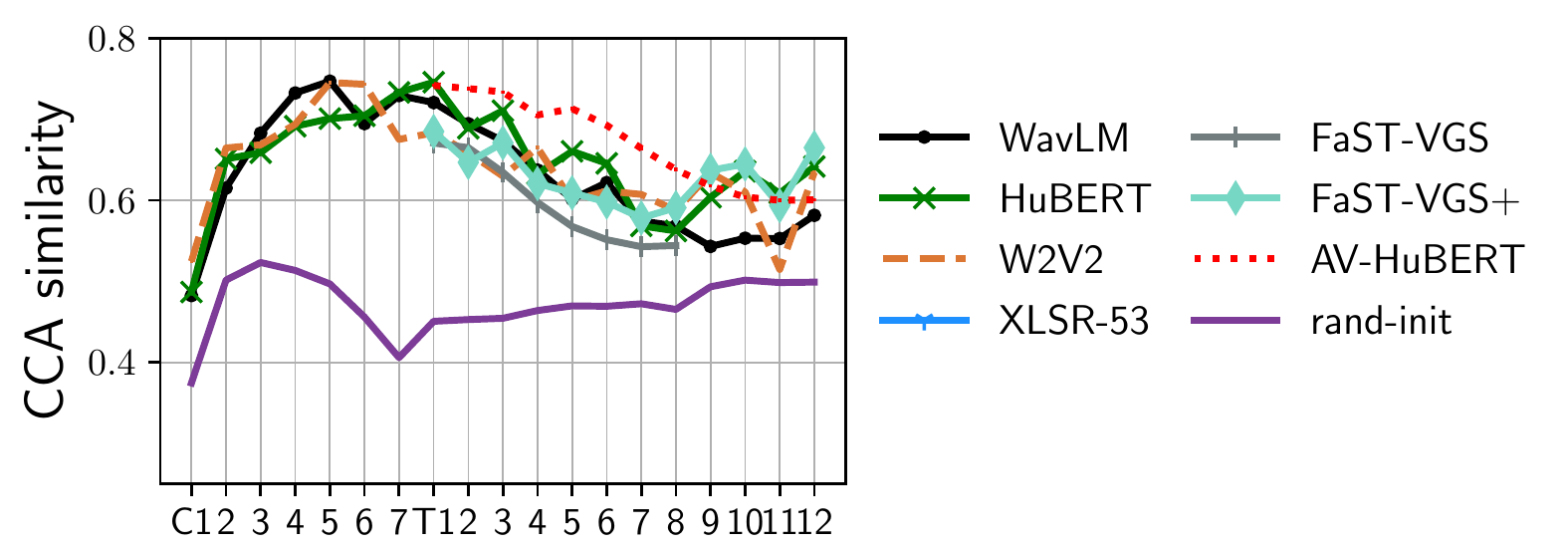}}
\end{minipage}
\begin{minipage}[b]{1.0\linewidth}

\vspace{-0.05cm}
\footnotesize
 \centering
 \centerline{\includegraphics[width=8cm, trim=0 0 0 157, clip]{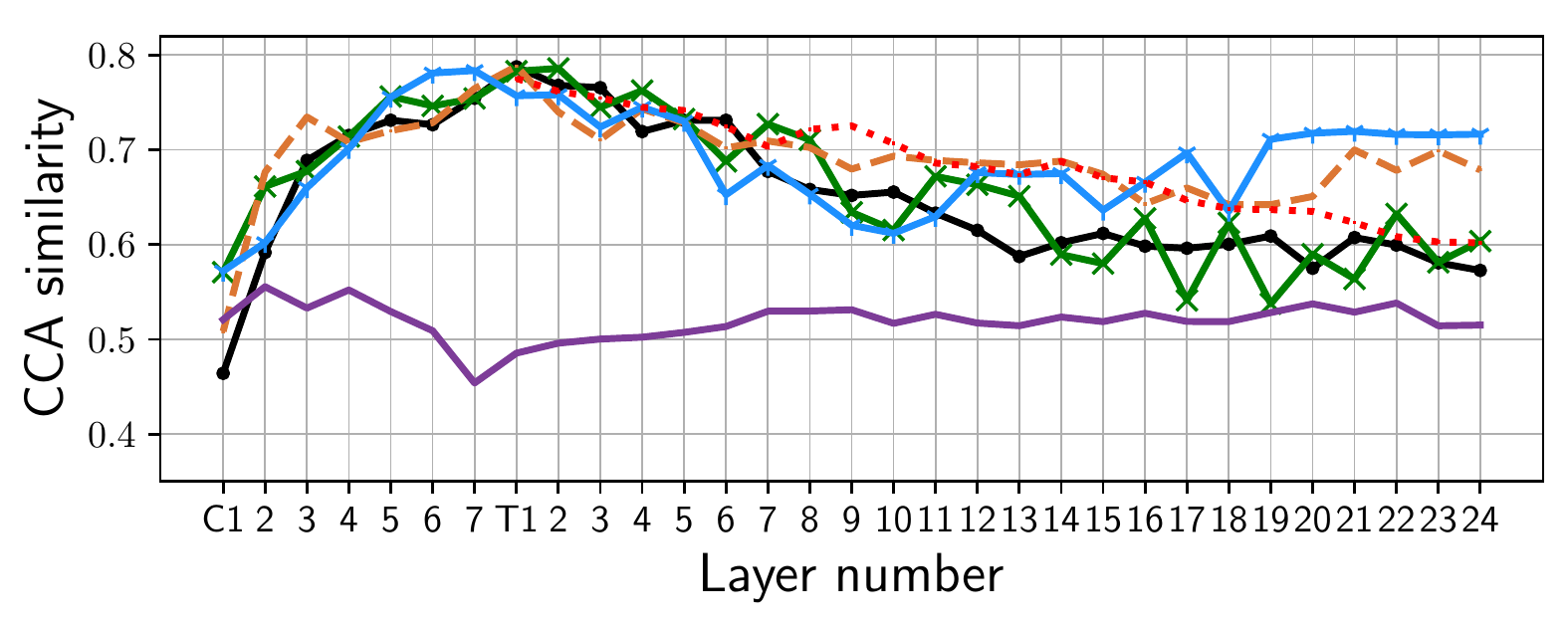}}
\end{minipage}

\vspace{-0.3cm}
\caption{\it CCA similarity with spectrogram features. $C_i$: CNN layer $i$, $T_j$: Transformer layer $j$.}
  \label{fig:cca-mel}
  \vspace{-0.5cm}
\end{figure}

{\bf Frame-level acoustic content.}
In \fig~\ref{fig:cca-mel}, we measure the correlation between pre-trained features and the widely used spectrogram (mel filter bank) features. For all models, the final CNN layers or the initial transformer layers are highly correlated with spectrogram features. Our findings are consistent with recent work using synthetic audio to analyze \wavtovec \ CNN layers ~\cite{choi2022opening}. This suggests a possible replacement of the CNN module with spectrogram features, and Wu et al.~\cite{wu2022performance} report comparable performance with this modification in some settings.

For the randomly initialized models, we see a non-trivial trend in the CNN layers, presumably because the CNN architecture has an inductive bias similar to the filtering mechanism of mel feature extraction. However, the correlation values are as low as the lowest scores for the pre-trained models. 
\vspace{.05in}

\begin{figure}[t]
\begin{minipage}[b]{1.0\linewidth}
\small

 \centering
 \centerline{\includegraphics[width=8cm, trim=0 184 0 0, clip]{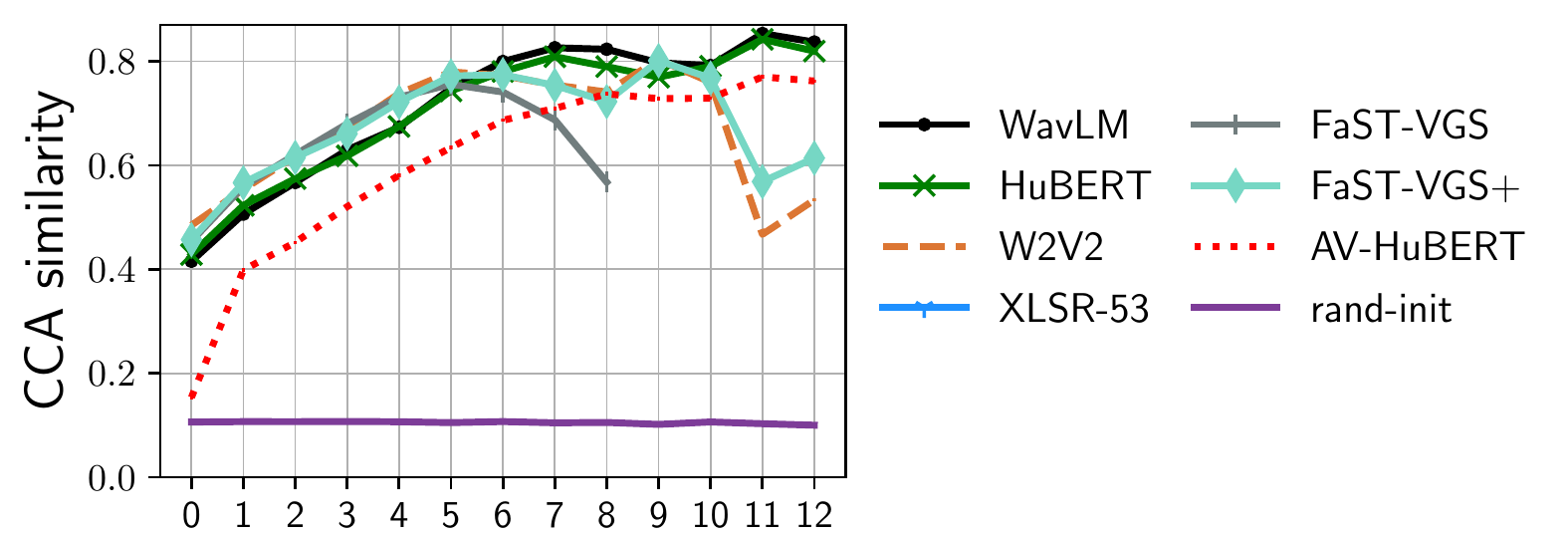}}
\end{minipage}
\begin{minipage}[b]{1.0\linewidth}

\vspace{-0.05cm}
\footnotesize
 \centering
 \centerline{\includegraphics[width=8cm, trim=0 0 0 155, clip]{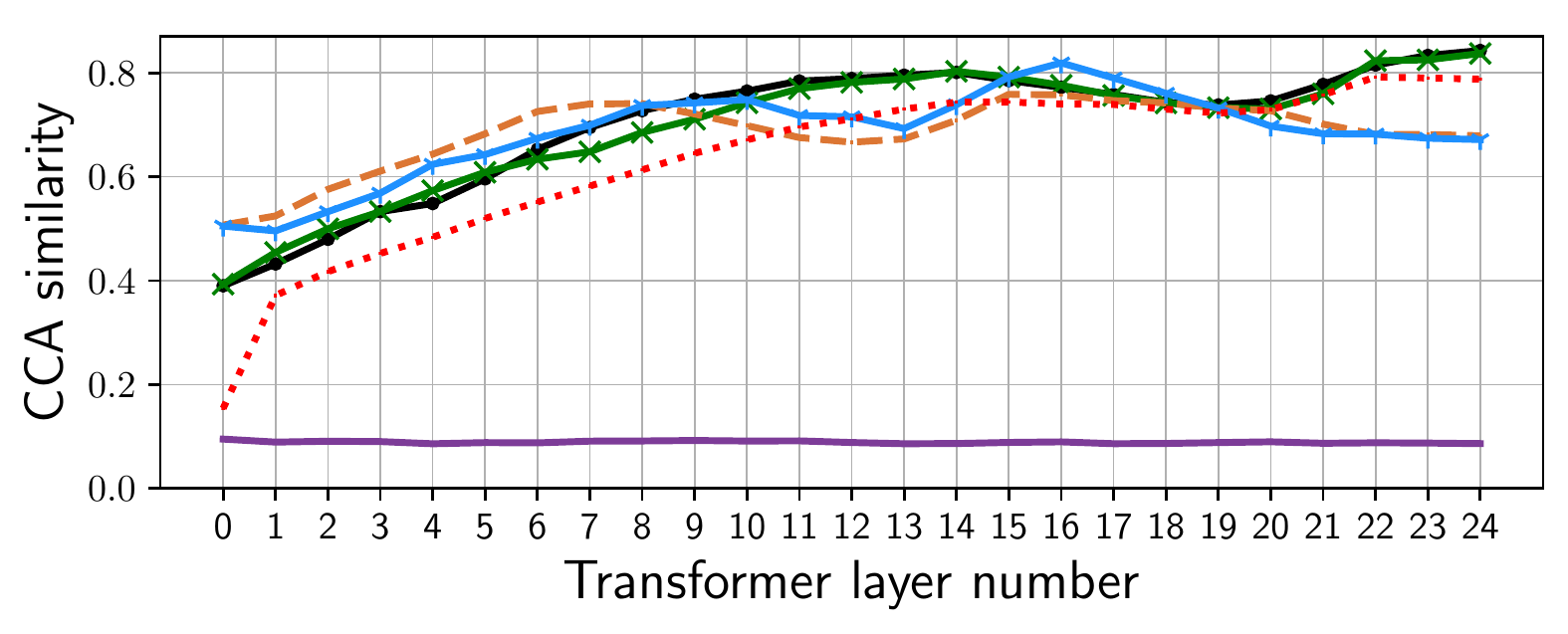}}
\end{minipage}

\vspace{-0.3cm}
\caption{\it CCA similarity with phone labels.}
  \label{fig:cca-phone}
\vspace{-0.3cm}
\end{figure}

\begin{figure}[t]
\begin{minipage}[b]{1.0\linewidth}
\small

 \centering
 \centerline{\includegraphics[width=8cm, trim=0 184 0 0, clip]{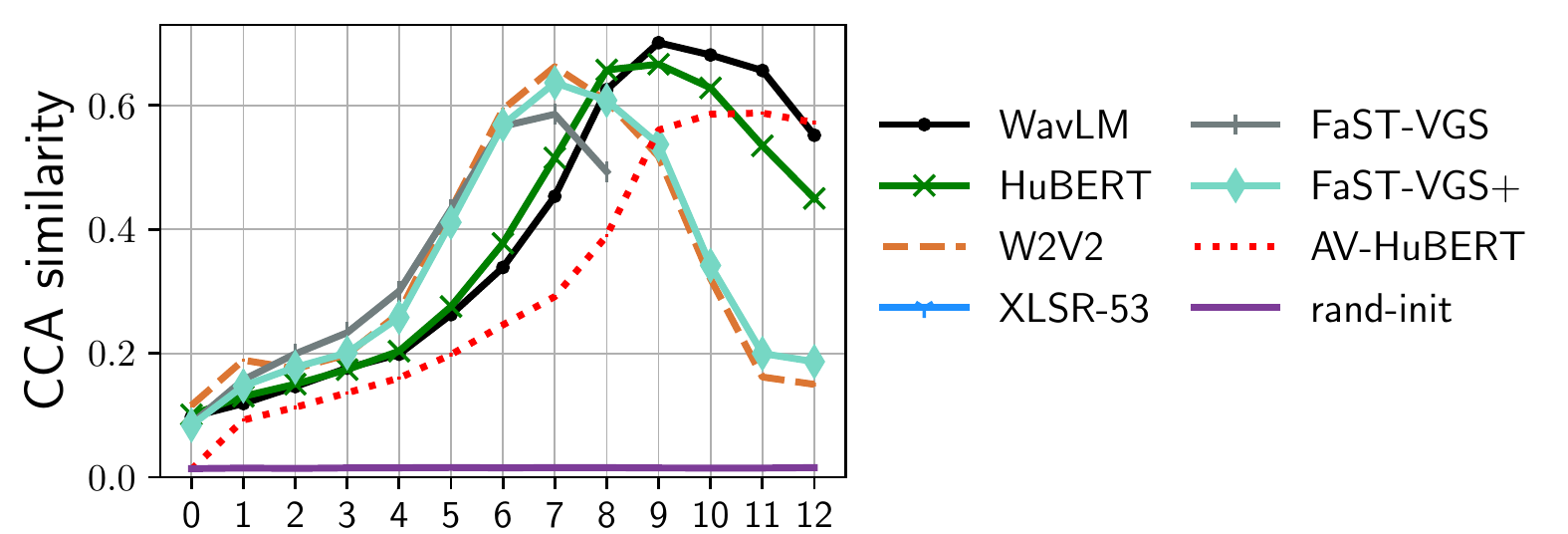}}
\end{minipage}
\begin{minipage}[b]{1.0\linewidth}

\vspace{-0.05cm}
\footnotesize
 \centering
 \centerline{\includegraphics[width=8cm, trim=0 0 0 155, clip]{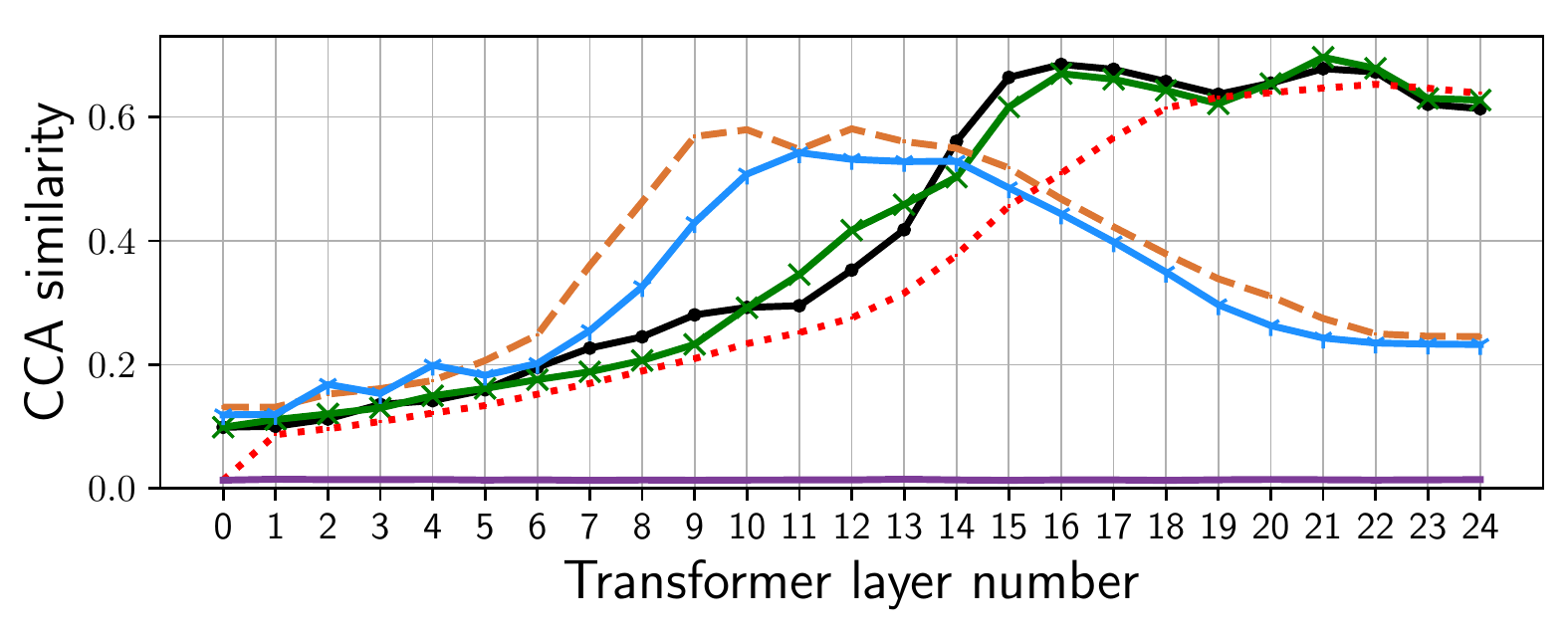}}
\end{minipage}

\vspace{-0.3cm}
\caption{\it CCA similarity with word labels.}
  \label{fig:cca-word}
  \vspace{-0.6cm}
\end{figure}

\noindent {\bf Phonetic and word-level information.}
\fig~\ref{fig:cca-phone} and \ref{fig:cca-word} show the layer-wise phonetic and word similarity respectively. Models that have a strong autoencoder-style dynamic (\wavtovec, \xlsr, and \fastvgsp, as seen in \fig \ref{fig:cca-intra}) tend to have a peak in both phonetic and word content in one or more of the intermediate layers, with a drop in the highest layers. These models have the same masking-based contrastive loss that recovers the local features. \fastvgsp's objective also includes a cross-modal loss. \fastvgs, on the other hand, is trained with only a cross-model loss, so the reason for \fastvgs's autoencoder-style behavior is unclear.

For the models which are trained to predict discrete units learned in an intermediate layer (\hubert, \wavlm, \avhubert) the phonetic and word information appears to be concentrated toward higher layers than for the other (``autoencoder-like") models. 
The finding for phonetic content is consistent with trends observed in \hubert \ measured via clustering~\cite{hsu2021hubert}. Hsu et al. also report that the first iteration of \hubert \ trained to predict discrete units learned from mel cepstral features (as opposed to an intermediate layer) has a peak phonetic content at a lower layer.
Such findings suggest that the trend may be more affected by the latent feature layer used in the pre-training objective rather than the form of the objective itself. 
In related work, analyses done by Chung et al.~\cite{chung2020similarity} suggest that self-supervised models are more affected by the training objective than the architecture; however, their work did not consider layer-wise trends or the multiple flavors of masked prediction objectives analyzed here.
Finally, our findings for word-level content in \hubert \ are consistent with a recent study analyzing acoustic word embeddings for word discrimination~\cite{sanabria2022analyzing}.

\vspace{-.12in}
\subsection{Implications for downstream tasks}
\label{sec:corr}
\begin{figure}[htb]
\footnotesize
 \centering
 \vspace{-0.3cm}
 \centerline{\includegraphics[width=8cm]{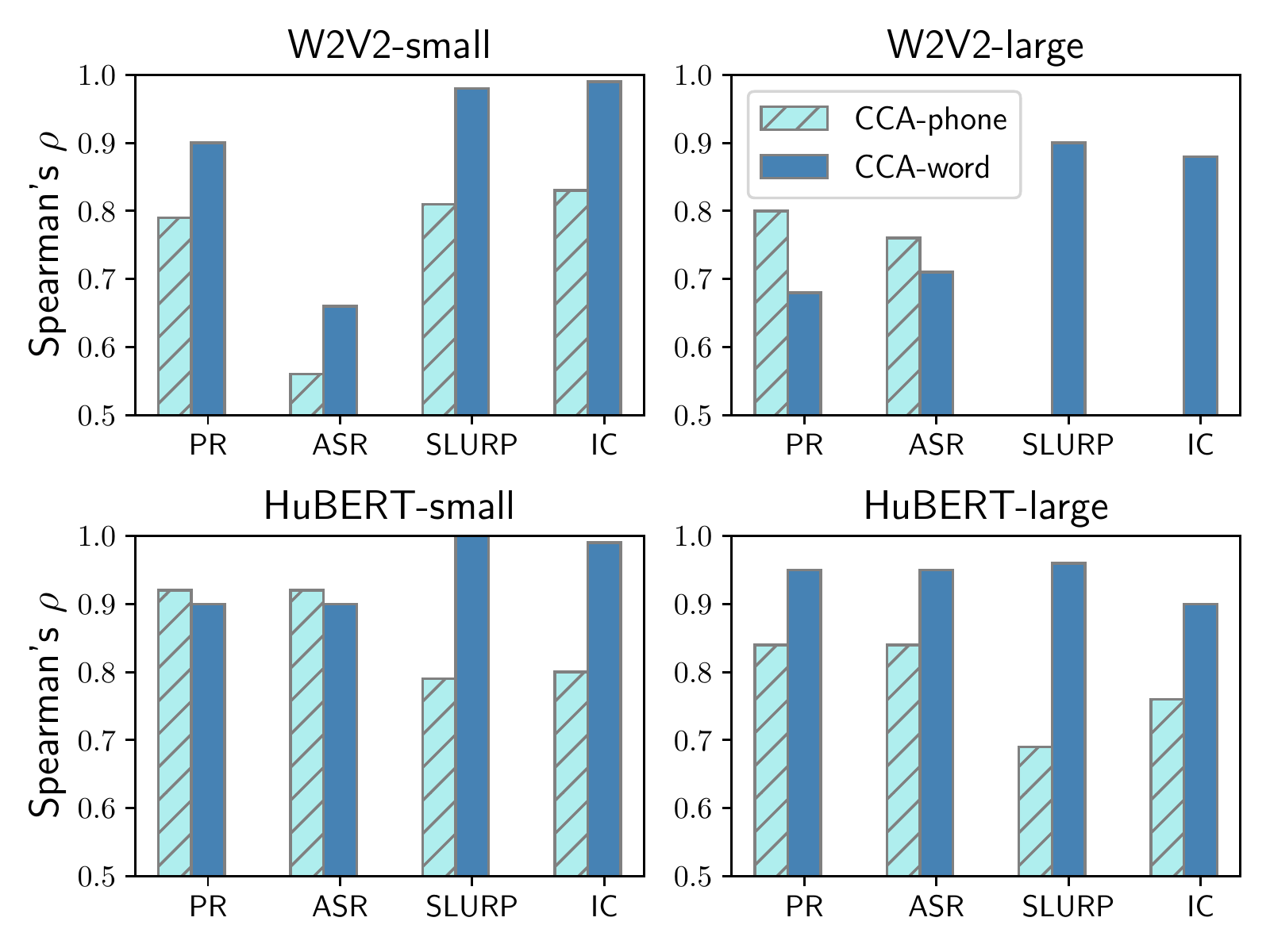}}
\vspace{-0.5cm}
\caption{\it 
Correlation between layer-wise CCA scores and task performance. ASR and PR performances are typically reported as error rates; for consistency we use ($100-\text{error rate}$) here, so a higher correlation is better. Scores $<0.5$ are not visible here. }
  \label{fig:corr-scores}
\vspace{-0.3cm}
\end{figure}

\vspace{-.1in}
Next, we study the relationship between layer-wise phone/word content and downstream task performance. \fig~\ref{fig:corr-scores} shows the correlation (specifically, Spearman's $\rho$ rank correlation, to account for potentially non-linear relationships) between layer-wise CCA scores and downstream performance for  \wavtovec \ and \hubert.\footnote{SLURP-action and SLURP-scenario have similar trends and takeaways (despite differences in absolute scores) so we present only SLURP-scenario.} As might be expected, PR and ASR performance are well-correlated with both CCA-phone and CCA-word scores. Semantic task (SLURP-action, SLURP-scenario, and IC) performance is much more correlated with CCA-word than with CCA-phone, presumably because these tasks benefit more from word information than phone information. 

\begin{figure}[t]
\footnotesize
 \centering
 \centerline{\includegraphics[width=9cm]{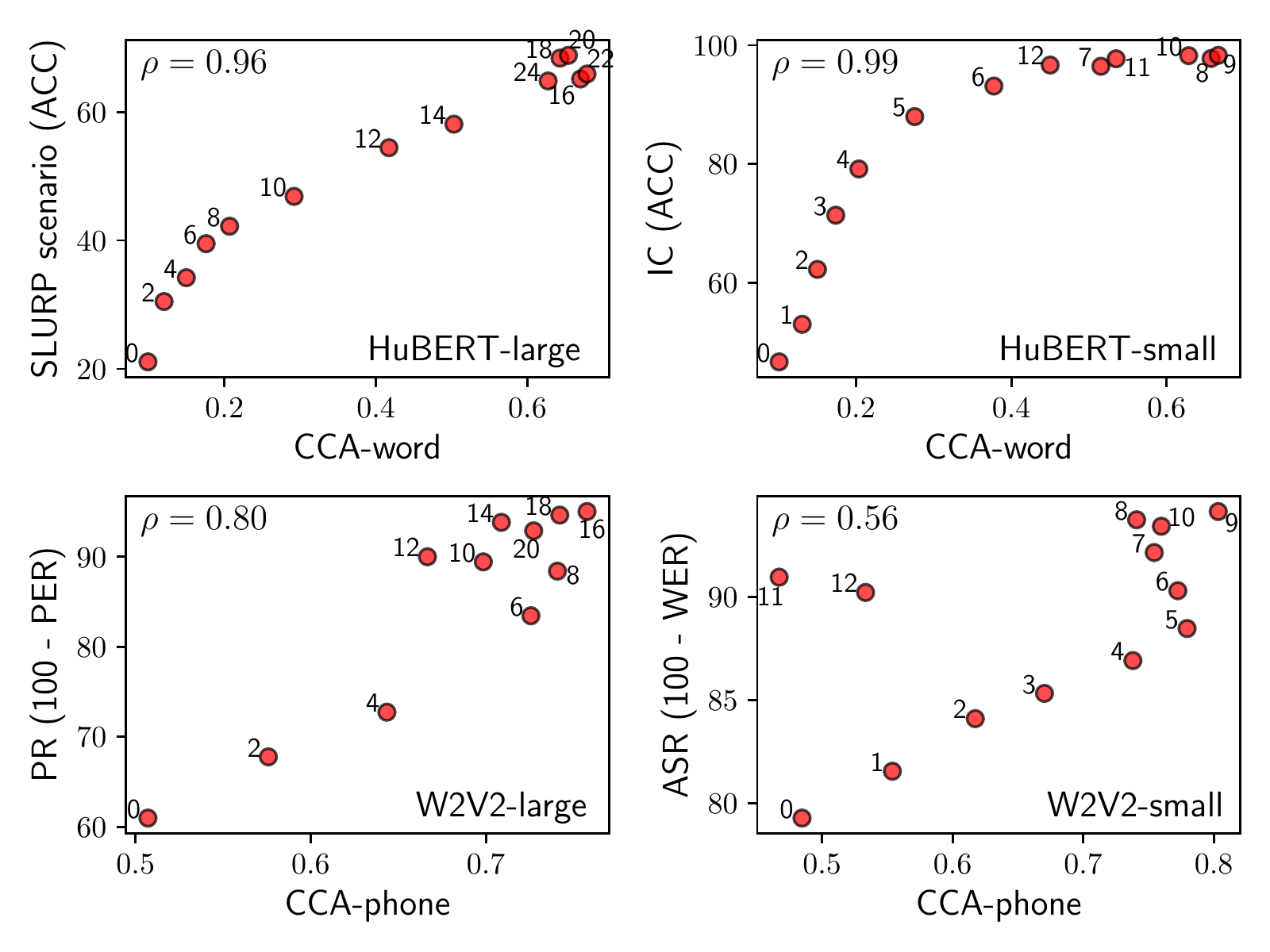}}
\vspace{-0.5cm}
\caption{\it Scatter plots for specific combinations of model, task, and CCA measure. Each point is labeled with its layer index.}
  \label{fig:scatter}
\vspace{-0.4cm}
\end{figure}

\fig~\ref{fig:scatter} shows scatter plots for four of the settings in \fig~\ref{fig:corr-scores}, including some of the low-scoring correlation combinations. We note that most of the top-performing layers are common for the CCA measure and the task performance, even though the Spearman's correlation does not capture this well in the presence of some outliers.

\begin{figure}[t]
\begin{minipage}[b]{1.0\linewidth}
\small

 \centering
 \centerline{\includegraphics[width=8cm, trim=0 305 0 10, clip]{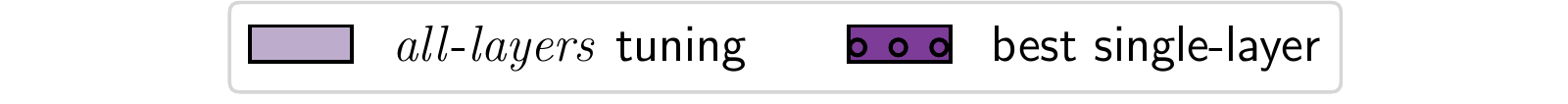}}
\end{minipage}
\begin{minipage}[b]{1\linewidth}

\vspace{-0.05cm}
\footnotesize
 \centering
 \centerline{\includegraphics[width=8cm]{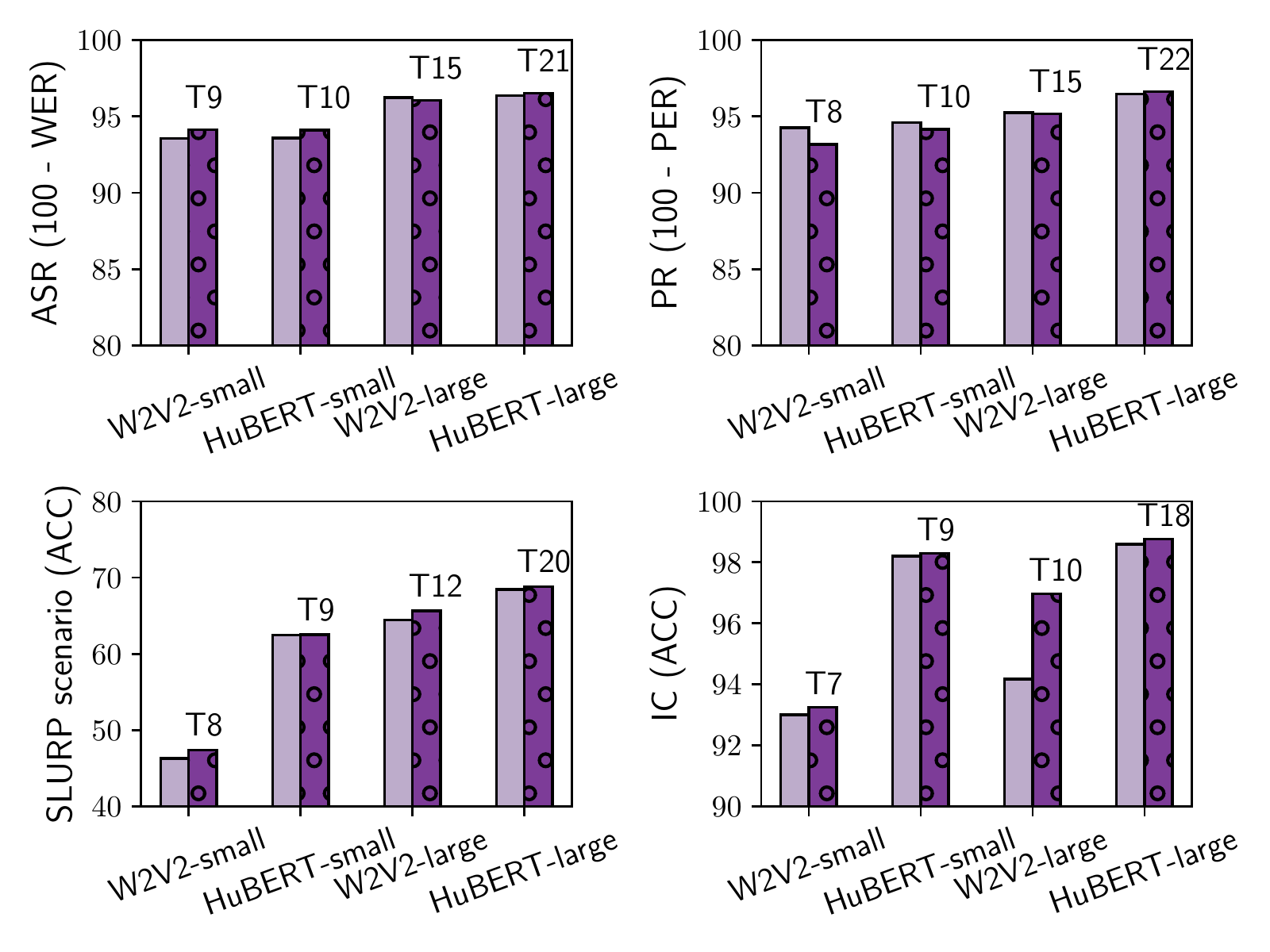}}
 \vspace{-0.15cm}
\end{minipage}

\caption{\it Task performance and best layer for all tasks.}
\label{fig:task-scores}
\vspace{-0.3cm}
\end{figure}

These findings imply that CCA similarity scores are indeed informative for downstream tasks. We compare the best layer performance to the task performance using a learnable weighted sum of all layers ({\tuning}), as in \cite{Yang2021SUPERBSP}. \fig~\ref{fig:task-scores} shows the scores from \tuning \ and the best single-layer performance, along with the best layer for the latter. We find that (i) the best single-layer ASR performance for \wavtovecs, \huberts, and \hubertl \ is better than that of \tuning \ (by 4-9\% relative), (ii) the best single-layer performance for semantic tasks matches or outperforms the \tuning \ performance, with the exception of SLURP-action with \hubert \ models (not shown in the figure),
(iii) the best-performing layer is always lower than at least the top two layers and is close to the layers observed to have the most phonetic and word-level content as measured by CCA (see \figs~\ref{fig:cca-phone} and \ref{fig:cca-word}). Specifically, for all semantic tasks, the best-performing layer is one of the top 3 in terms of CCA-word; and for ASR and PR, the best-performing layer is one of the top 6 in terms of both CCA-phone and CCA-word. 

It is evident that a single intermediate layer often matches the \tuning \ performance. But evaluating each individual layer to find the best-performing layer is much more computationally demanding than evaluating CCA, which requires fewer samples and has a closed-form solution.
Our findings suggest using CCA-word/CCA-phone to narrow down the choice of layers, thus reducing compute requirements for task-specific adaptation. Furthermore, using an intermediate layer also reduces space and compute resources as we discard the top few layers. Similar results have been found for text models when applying light-weight analysis tools to guide efficient fine-tuning~\cite{xie2022hidden}.

We note that although analyses like ours are often indicative of downstream performance, task performance depends not only on the representation but also on modeling and optimization issues.  For example,
layers 22 and 24 of \wavtovecl \ do not converge on speech recognition tasks despite having a decent CCA-phone score of 0.6.

Alternatively to the presented approach, it is natural to ask whether the layer weights learned in \tuning \ experiments are themselves a good indicator of usefulness for downstream tasks.  In our experiments, the mean rank correlation between layer weights and task performance is 0.66 across all ten (task, model) pairs.  In contrast, the mean rank correlation between CCA-word and task performance is 0.90, even though CCA-word is a generic measure while layer weights are task-specific.  The relative unreliability of layer weights as an analysis tool is not too surprising and is similar to findings for attention weights in NLP models~\cite{jain2019attention, wiegreffe2019attention}.

\vspace{-0.1in}
\section{Conclusion}
\vspace{-0.1in}
The analyses presented here are, to our knowledge, the largest cross-model layer-wise comparisons for pre-trained speech representations.  Our findings suggest that useful phone/word information tends to concentrate in different layers depending on the model (and specifically its pre-training objective) and that these findings can help guide application of the models for downstream tasks. It is encouraging to see that multiple studies in the recent literature, using different analysis tools, point to some similar conclusions.

We are able to derive implications for a variety of tasks by analysing a few fundamental acoustic-linguistic properties. Since our analytical method is task-agnostic, it is more scalable than popular task-specific probing approaches for layer-wise analysis~\cite{shah2021all, feng2022silence, ji2022predicting}. Our analysis techniques are easily extensible to additional properties besides the ones we have studied here (e.g., speaker, prosody, syntax).  We make our code publicly available in the hope that it will enable additional large-scale analyses, and help to navigate the increasingly large landscape of representation models.  

\bibliography{refs}
\bibliographystyle{IEEEbib}

\end{document}